\documentclass{article}
\usepackage{iclr2027_conference,times}
\iclrfinalcopy
\usepackage[utf8]{inputenc}
\usepackage[T1]{fontenc}
\usepackage{amsmath,amssymb,amsthm,booktabs,graphicx,tabularx}
\usepackage{microtype}
\usepackage{hyperref,url}
\hypersetup{colorlinks=true,linkcolor=black,citecolor=blue,urlcolor=blue,pdftitle={Monarch-MoE: Structured Experts for Multimodal Continual Instruction Tuning},pdfauthor={Xi Zhang, Boyu Diao, Zesen Chen, Ruiqi Liu, Hangda Liu, Yongjun Xu}}

\newcommand{\rsf}{\mathrm{F}_1}
\title{Monarch-MoE: Structured Experts for\\Multimodal Continual Instruction Tuning}
\author{%
\begin{tabular}[t]{c}
Xi Zhang$^{1,2}$ \quad
Boyu Diao$^{1,2}$\thanks{Corresponding author: \href{mailto:diaoboyu2012@ict.ac.cn}{\texttt{diaoboyu2012@ict.ac.cn}}.} \quad
Zesen Chen$^{3}$ \\[0.3em]
Ruiqi Liu$^{1,2}$ \quad Hangda Liu$^{1,2}$ \quad Yongjun Xu$^{1,2}$ \\[0.6em]
\normalfont\small $^1$Institute of Computing Technology, Chinese Academy of Sciences \\
\normalfont\small $^2$University of Chinese Academy of Sciences \\
\normalfont\small $^3$ShanghaiTech University
\end{tabular}}
\begin{document}
\maketitle
% Public preprint: remove the conference acceptance/review header.
\fancyhead{}
\renewcommand{\headrulewidth}{0pt}
% ABSTRACT accepted and frozen by the author on 2026-09-26 (R018 / v011).
% Exact current baseline: abstract.lock.json; earlier baselines remain in version history.
\begin{abstract}
Learning new tasks in multimodal continual instruction tuning can cause catastrophic forgetting of previously acquired knowledge.
Although multi-expert LoRA distributes learning across experts, updates to a dense low-rank expert can still affect many output features.
We introduce Monarch-MoE, which replaces the dense factors in LoRA experts with block-diagonal factors connected by a fixed permutation.
This structure provides blockwise gradient isolation: individual block-pair updates affect only one output block, while connections between all input and output blocks are maintained.
To make these compact experts practical on memory-limited hardware, we combine batched computation, fused forward operators, and an optimized backward pass.
Across three continual-learning methods evaluated on MLLM-DCL and UCIT with LLaVA-1.5-7B, Monarch-MoE improves first-task retention while reducing adapter parameters by approximately 75\% and 50\%, respectively.
In the main comparison, first-task forgetting is as low as 0.52 percentage points on MLLM-DCL and 0.10 on UCIT, with method-dependent trade-offs in final performance.
Our implementation supports fine-tuning and inference on a single 24\,GB RTX~4090, with lower measured memory use and competitive runtime in the evaluated settings.
Together, these algorithmic and systems improvements make structured experts a practical option for continual adaptation on memory-limited hardware.
\end{abstract}
% INTRO-D05 accepted and frozen by the author on 2026-09-26 (R018 / v011).
% Exact text and Figure 1 asset hashes: introduction.lock.json.
\section{Introduction}
\label{sec:intro}

% INTRO-D02-P01: capabilities -> evolving applications -> sequential adaptation -> forgetting
Large vision-language models have demonstrated strong capabilities across a broad range of visual understanding and instruction-following tasks~\citep{liu2024llava15}.
In practical applications, however, the demands placed on these models can evolve: an assistant trained for medical visual question answering may later need to interpret documents without losing its earlier capabilities.
Learning across such task sequences while retaining earlier skills is the objective of multimodal continual instruction tuning~\citep{gao2024moelora}.
During adaptation to each new task, training examples from earlier tasks are unavailable~\citep{guo2024mcitlib}.
Here, catastrophic forgetting becomes a central challenge: updates driven by new data can alter parameters on which earlier tasks depend, degrading previously acquired capabilities~\citep{kirkpatrick2017ewc}.

% INTRO-D02-P02: repeated adaptation cost -> PEFT -> shared adapter -> multiple experts
Repeatedly updating all parameters of these large models is costly.
Parameter-efficient fine-tuning (PEFT) freezes the backbone and learns compact adaptation modules; LoRA represents each weight update as a low-rank product $BA$~\citep{hu2022lora}.
When a single adapter is reused across a task sequence, successive tasks modify the same trainable factors.
Multi-expert approaches offer greater flexibility by distributing adaptation across several modules; MoELoRA, for example, combines multiple low-rank experts through learned input-dependent gates~\citep{gao2024moelora}.
However, maintaining multiple trainable experts and their optimizer states can still place substantial demands on a single GPU.

% INTRO-D02-P03: expert allocation -> within-expert sharing -> structural motivation
Choosing an expert does not determine how localized an update is within it.
In a LoRA expert $BA$, changing one entry of $A$ with $B$ fixed can affect many output coordinates through the corresponding column of $B$.
When an expert is reused across tasks, this shared dependence gives new-task updates a path to affect features used by earlier tasks.
This motivates structuring each expert so that individual updates act locally while preserving communication across the feature space.

\begin{figure}[t]
\centering
\includegraphics[width=\linewidth]{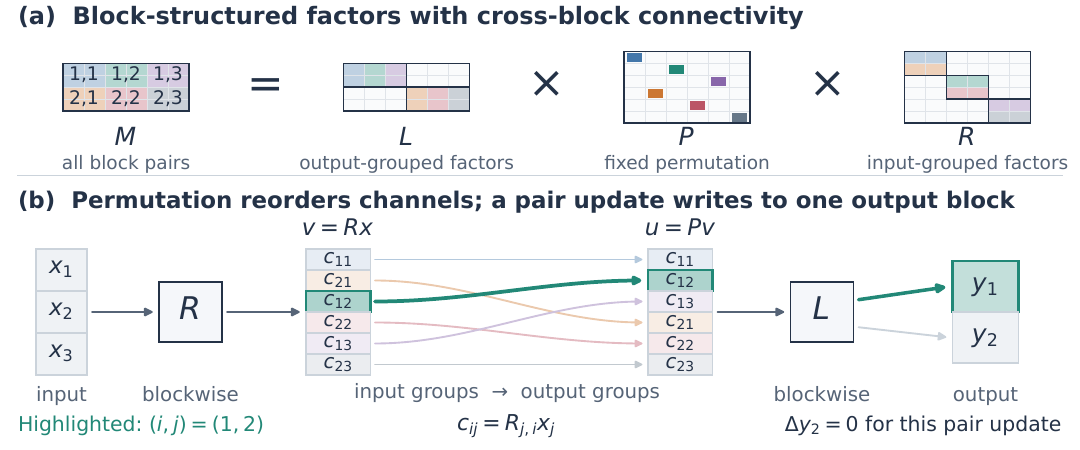}
\caption{\textbf{Monarch factorization and channel permutation.} (a) $M=LPR$ connects all input--output block pairs. (b) $P$ reorders the channels. The highlighted pair update changes only $y_1$ at the expert output; full steps can update all pairs. Colors denote block pairs, not tasks.}
\label{fig:structure}
\end{figure}

% INTRO-D02-P04: structured factors -> local channel updates + cross-block connectivity -> integration
We introduce \textbf{Monarch-MoE}, which replaces the dense low-rank factors within experts with the structured parameterization illustrated in Figure~\ref{fig:structure}.
Building on Monarch matrices~\citep{dao2022monarch} and their rectangular formulation for parameter-efficient fine-tuning~\citep{tan2024more}, we construct each expert from two block-diagonal factors connected by a fixed permutation.
The construction partitions input and output features into blocks and assigns a separate low-rank channel to every input--output block pair.
Each channel receives gradients through its associated output block and writes updates only to that block, while the permutation preserves connections between all input and output blocks.
We call this property blockwise gradient isolation.
We hypothesize that it favors task updates in less-overlapping subspaces, consistent with the near-orthogonal update subspaces observed in our analysis.
The structured experts can be integrated into existing continual-learning methods while retaining their learning strategies.

% INTRO-D03-P05: compact parameters -> execution overhead -> training/inference implementation
Making structured experts practical also requires efficient execution.
A compact factorization can incur substantial overhead when evaluated through many small matrix operations and intermediate rearrangements.
We consolidate the expert factors for batched computation, fuse the $R\!\to\!P\!\to\!L$ forward computation for supported shapes, and implement an explicit batched backward pass.
For autoregressive inference, model-level compilation further reduces dispatch overhead.
Together with the smaller adapter and optimizer-state footprint, these optimizations support fine-tuning and inference on a single 24\,GB RTX~4090.

% INTRO-D03-P06: algorithmic evidence and practical feasibility, scoped to evaluated settings
We evaluate the algorithmic and practical effects separately using LLaVA-1.5-7B and the MCITlib protocols~\citep{liu2024llava15,guo2024mcitlib}.
Continual-learning experiments show improved first-task retention with fewer adapter parameters across several learning strategies, while also exposing trade-offs in new-task acquisition.
Capacity controls, repeated runs, and representation analyses examine how expert structure relates to this behavior.
Single-GPU measurements assess training throughput, inference cost, and memory use, establishing practical feasibility for the evaluated configurations.

% INTRO-D03-P07: two contributions; empirical analyses support the algorithmic contribution
Our contributions are twofold:
\begin{itemize}
  \item \textbf{Structured experts for mitigating update interference.} We develop Monarch-MoE with blockwise gradient isolation and cross-block connectivity, and analyze its local update support and learned representations. Experiments show strong continual-learning performance and improved retention with fewer adapter parameters.
  \item \textbf{Memory-efficient execution on a single GPU.} We combine compact expert factors with batched computation, fused forward operators, and optimized backward and inference execution. The resulting implementation supports fine-tuning and inference on a single 24\,GB RTX~4090, with competitive runtime in the evaluated settings.
\end{itemize}

% RELATED-D01 / R019: revised under the author's instruction; pending review.
\section{Related Work}
\label{sec:related}
\textbf{Multimodal continual instruction tuning.}
CoIN evaluates learning and forgetting across sequences of visual instruction tasks~\citep{gao2024moelora}; MCITlib unifies multiple benchmarks and continual-learning implementations~\citep{guo2024mcitlib}.
Complementing these evaluation efforts, HiDe-LLaVA uses layerwise representation analysis to guide task-specific expansion and task-general fusion~\citep{guo2025hide}.
We follow MCITlib's setting without replaying previous-task training examples and examine expert structure within its continual-learning implementations.

\textbf{Parameter-efficient continual learning.}
LoRA and DoRA learn compact updates to pretrained weights~\citep{hu2022lora,liu2024dora}; continual learning additionally requires managing these updates across tasks.
Existing strategies use subspace constraints, expert routing, or knowledge organization.
O-LoRA freezes previous adapters and penalizes overlap between new and previous low-rank factors~\citep{wang2023olora}.
MoELoRA distributes adaptation across experts with input-dependent gates~\citep{gao2024moelora}, and CL-MoE combines instance- and task-level routing with momentum updates for continual visual question answering~\citep{huai2025clmoe}.
In federated continual instruction tuning, DISCO organizes task knowledge in subspaces and selectively activates their outputs~\citep{zhang2024disco}.
Monarch-MoE complements these strategies by restricting the direct output and gradient support of block-pair updates inside each expert.
This structure adds no cross-task orthogonality penalty; replacement experiments retain the host method's constraints.
Near-orthogonal learned update subspaces are an empirical observation, not a structural guarantee.

\textbf{Structured adaptation and efficient execution.}
Monarch combines block-diagonal factors with permutations and supports hardware-efficient batched matrix multiplication~\citep{dao2022monarch}.
MoRe applies rectangular Monarch factors to parameter-efficient fine-tuning~\citep{tan2024more}.
Building on these foundations, we study structured experts for continual learning and develop their multi-expert implementation through batched expert computation, fused forward operators, and an optimized backward pass.
Training and inference measurements on a single RTX~4090 assess memory requirements and runtime in the tested configurations.

% METHOD-D01 / R022: algorithm-first revision under the author's instruction.
\section{Monarch Experts for Continual Adaptation}
\label{sec:method}
\subsection{Problem Setup and Adaptation Framework}
We consider task datasets $\mathcal D_1,\ldots,\mathcal D_T$ arriving sequentially. At stage $t$, the model learns from $\mathcal D_t$ without replaying earlier training examples and is evaluated on tasks $1,\ldots,t$. A frozen backbone is adapted through trainable expert factors and the components required by the host continual-learning method.

For a frozen weight $W\in\mathbb R^{d_o\times d_i}$ and input feature $x$, the adapted layer is
\begin{equation}
 z=Wx+\sum_{e=1}^{K}g_e(x)\,s_eM_ex.
 \label{eq:moe}
\end{equation}
Here $M_e$ is an expert update matrix, $g_e$ its dispatch weight, and $s_e$ its output scale. Standalone models use top-2 dispatch; replacements follow the host method's routing or task-indexed selection, so $g_e$ may also depend on task information. Monarch-MoE changes the factors defining $M_e$. We next describe how this changes the dependence between expert parameters and output features.

\subsection{Monarch-Parameterized Experts}
We build on Monarch matrices~\citep{dao2022monarch} and their rectangular use in parameter-efficient adaptation~\citep{tan2024more}. Split the input into $m_w$ blocks $x_j\in\mathbb R^w$ and the output into $m_h$ blocks of size $h$, where $w=d_i/m_w$ and $h=d_o/m_h$. Omitting the expert index, define
\begin{equation}
 R_{j,i}\in\mathbb R^{r\times w},\qquad
 L_{i,j}\in\mathbb R^{h\times r},\qquad
 (Mx)_i=\sum_{j=1}^{m_w}L_{i,j}R_{j,i}x_j.
 \label{eq:pairs}
\end{equation}
Each input--output block pair has its own channel of inner dimension $r$, and its product has rank at most $r$.

Stacking $R_{j,i}$ over $i$ within each input block forms block-diagonal $R$; concatenating $L_{i,j}$ over $j$ within each output block forms block-diagonal $L$. A fixed permutation $P$ reorders the intermediate coordinates from $(j,i,a)$ to $(i,j,a)$, $a=1,\ldots,r$, giving $M=LPR$ (Figure~\ref{fig:structure}). Every input block can therefore reach every output block, while the factors assigned to distinct block pairs remain separate. The partition concerns hidden feature coordinates; no task-to-block assignment is prescribed. The permutation supplies cross-block connectivity without trainable parameters. For symmetric partitions, setting $P=I$ instead gives a block-diagonal map, although its global rank need not decrease (Appendix~\ref{app:gradient}).

\subsection{Local Update Support and Cross-Task Interference}
\label{sec:local_support}
An expert's factorization determines which output coordinates depend on a parameter change. In a dense LoRA expert $M=BA$~\citep{hu2022lora}, a change to $A$ with $B$ fixed produces $B\Delta Ax$. Changing one entry of $A$ can affect all output coordinates where the corresponding column of $B$ is nonzero. Thus a shared input factor provides a path for new-task updates to alter features used by earlier tasks. We examine this dependence locally, holding the input, dispatch, and other factors fixed.

In Monarch, each pair has a more restricted update support. Let $E_i$ embed output block $i$ into the full vector and $S_j$ select input block $j$, so that $M=\sum_{i,j}E_iL_{i,j}R_{j,i}S_j$. If only one pair changes, its exact finite update is
\begin{equation}
 \Delta M=E_i\bigl(\Delta L_{i,j}R_{j,i}+L_{i,j}\Delta R_{j,i}
                  +\Delta L_{i,j}\Delta R_{j,i}\bigr)S_j.
 \label{eq:localchange}
\end{equation}
Hence $E_{i'}^\top\Delta M=0$ for $i'\ne i$: the update directly writes to one output block. The highlighted channel in Figure~\ref{fig:structure} illustrates this property, including when both factors change.

The same organization governs gradient dependence. For the unscaled expert output $y=Mx$, define $\delta_i=E_i^\top\nabla_y\mathcal L$. This gradient already includes the chain-rule contribution of dispatch and output scale in Eq.~\eqref{eq:moe}. Then
\begin{equation}
 \frac{\partial\mathcal L}{\partial L_{i,j}}=\delta_i(R_{j,i}x_j)^\top,
 \qquad
 \frac{\partial\mathcal L}{\partial R_{j,i}}=L_{i,j}^\top\delta_i x_j^\top.
 \label{eq:grad}
\end{equation}
Both factors receive gradients directly through output block $i$. We call this blockwise gradient isolation. The permutation maintains input--output connectivity while the factorization limits this direct sharing across output blocks; the upstream gradient itself can still depend on the rest of the network.

This distinction motivates our retention hypothesis. Dense input factors couple multiple output coordinates through shared parameters; separate block-pair factors allow an update to one output block without changing the factors feeding other output blocks. We hypothesize that this organization favors task updates in less-overlapping subspaces and helps preserve earlier representations. Any task-dependent specialization must emerge during optimization. Local support does not assign different tasks to disjoint blocks: a full optimizer step may update every pair, tasks may reuse pairs, and routing or subsequent layers can introduce further interactions.

We evaluate this hypothesis through complementary analyses. Capacity and partition controls examine alternative explanations for retention; old-task representation CKA measures stability; principal angles between learned update subspaces characterize directional overlap. Expert-utilization summaries additionally examine whether gross routing imbalance explains the comparison. These analyses concern learned representations and updates, not a guarantee that all task gradients are orthogonal. Their results and scope appear in Section~\ref{sec:exp} and Appendix~\ref{app:mechanism}; Appendix~\ref{app:gradient} details the local property.

\subsection{Parameter Budget and Training}
Summing the factors over block pairs gives
\begin{equation}
 N_{\mathrm{factor}}=r(m_wd_o+m_hd_i),\qquad
 \rho=m_hm_wr,\qquad
 \operatorname{rank}(M)\leq\min(d_i,d_o,\rho).
 \label{eq:count}
\end{equation}
For $m_h=m_w=m$ and $d_i=d_o=d$, the count is $2dmr$ and the rank bound $m^2r$ is attainable when $m^2r\leq d$ (Appendix~\ref{app:rank}). At the same budget, dense LoRA has rank at most $mr$; matching Monarch's rank bound requires $m$ times as many factor parameters. This compares budgets and attainable ranks, not identical matrix families or guaranteed performance.

We distinguish per-expert LoRA rank $q$, the codebase's total-rank budget $Kq$, and Monarch's inner dimension $\rho$. Implemented scales are $s_{\mathrm M}=\alpha/(m_hr)$ and, for the standalone LoRA baseline, $s_{\mathrm L}=\alpha_{\mathrm L}/q$. The Monarch denominator is not $\rho$. Parameter- or rank-matching therefore does not ensure matched update scale; Appendix~\ref{app:config} gives configurations and comparison limits.

By default, $L$ is zero-initialized and $R$ is random, giving a zero adapter residual. Training uses the instruction-tuning loss with host-specific constraints where applicable. In the O-LoRA replacement, the host's orthogonality strategy is adapted to blockwise overlaps between current and previous $R$ factors, with explicit normalization (Appendix~\ref{app:integration}). This penalty belongs to the host integration, not the structural property above. Optional blockwise SVD initialization and its interaction with sparse gates are described in Appendix~\ref{app:svd}.

\subsection{Memory-Efficient Training and Inference}
\label{sec:execution}
Compact factors also reduce gradient and optimizer-state storage. To limit execution overhead, we batch expert computations and produce intermediate channels directly in output-block order, avoiding a separate permutation copy. The training path uses a custom backward that recomputes these channels from saved inputs and factors, reducing intermediate storage at the cost of recomputation.

For supported shapes, a fused forward operator computes $R\!\to\!P\!\to\!L$ with on-chip intermediates; a batched path handles other configurations. Model-level compilation further reduces dispatch overhead during generation. Sparse top-2 gating need not skip unselected expert arithmetic, and operator-level improvements need not translate into faster generation. Section~\ref{sec:systems} evaluates training and inference on a single RTX~4090 separately; Appendix~\ref{app:efficiency} provides implementation details and measurement scope.

% EXP-D01 / v014 / R024: approved EXP-O01 implemented; new prose pending author review.
\section{Experiments}
\label{sec:exp}
We evaluate whether structured experts improve continual learning, examine alternative explanations based on capacity and training choices, and connect the observed behavior to the update mechanism in Section~\ref{sec:local_support}. We then measure the cost of training and inference on a single GPU.

\subsection{Experimental Setup}
\textbf{Tasks and protocol.}
We use LLaVA-1.5-7B~\citep{liu2024llava15} with the MCITlib protocols~\citep{guo2024mcitlib}. MLLM-DCL evaluates five successive domains, while UCIT spans six tasks including classification, question answering, and captioning. Training uses only the current task's examples; earlier training data are not replayed. Adapters target attention and MLP projections. Task orders, learning rates, epochs, and implementation configurations appear in Appendix~\ref{app:config}.

\textbf{Comparison groups.}
Table~\ref{tab:main} includes existing continual-learning methods, standalone LoRA/Monarch mixtures, and Monarch replacements within CL-MoE, O-LoRA, and DISCO. MCITlib baselines come from its published LLaVA protocol and archived task matrices; we do not rerun them here. The standalone DCL pair matches routing and the exact adapter budget of 164,462,592 parameters. The UCIT pair instead uses approximately 246.7M LoRA and 126.8M Monarch parameters, and host replacements also compare different budgets. Counts exclude the frozen backbone and non-adapter tensors. Full-sequence results are individual runs; the three-seed study repeats only the first two DCL tasks.

\textbf{Evaluation measures.}
We assess new-task learning together with retention. MFT averages scores immediately after learning each task; MFN averages final scores after the sequence; MAA averages performance over learned tasks across stages. BWT measures the mean change from immediate to final performance on earlier tasks. First-task forgetting $\rsf$ measures the first task's score decrease from its initial checkpoint, rather than average forgetting across tasks: it concerns RS on DCL and ImageNet-R on UCIT. Accuracy-like scores use a 0--100 scale; captioning retains MCITlib's aggregate scores. Appendix~\ref{app:metrics} defines the metrics and rounding conventions.

\subsection{Continual Learning Performance}
\begin{table*}[t]
\centering
\caption{\textbf{Full continual-learning results on MLLM-DCL and UCIT.} $\dagger$: MCITlib baselines. \textbf{Bold}: best value per metric and benchmark among the compared methods. Parameters include adapter factors and routers (millions); $\sim$ denotes a configuration estimate. Only the DCL standalone pair is parameter matched. Drop-in $\alpha$: CL-MoE/O-LoRA 16 on DCL and 64 on UCIT; DISCO 256 on both. See Appendix~\ref{app:config} for counts and Appendix~\ref{app:metrics} for metric definitions.}
\label{tab:main}
\begingroup
\fontsize{8}{10}\selectfont
\setlength{\tabcolsep}{2.2pt}
\begin{tabular}{@{}l r ccccc r ccccc@{}}
\toprule
& \multicolumn{6}{c}{\textbf{MLLM-DCL (5 tasks)}} & \multicolumn{6}{c}{\textbf{UCIT (6 tasks)}}\\
\cmidrule(lr){2-7}\cmidrule(lr){8-13}
Method & Params & MFT$\uparrow$ & MFN$\uparrow$ & MAA$\uparrow$ & BWT$\uparrow$ & $\rsf\downarrow$ & Params & MFT$\uparrow$ & MFN$\uparrow$ & MAA$\uparrow$ & BWT$\uparrow$ & $\rsf\downarrow$\\
\midrule
\multicolumn{13}{l}{\textit{Existing methods (LoRA experts)}}\\
LoRA-FT$^{\dagger}$ & $\sim$80 & 64.98 & 53.00 & 61.13 & -14.97 & 8.67 & $\sim$40 & \textbf{76.89} & 61.24 & 76.55 & -18.78 & 33.97 \\
MoELoRA$^{\dagger}$ & $\sim$93 & 64.94 & 54.78 & 61.76 & -12.71 & 1.55 & $\sim$52 & 71.17 & 58.98 & 75.08 & -14.63 & 21.16 \\
CL-MoE$^{\dagger}$ & $\sim$405 & 66.06 & 54.88 & 61.79 & -13.97 & 7.74 & $\sim$247 & 71.46 & 58.49 & 74.19 & -15.56 & 24.97 \\
O-LoRA$^{\dagger}$ & $\sim$400 & 65.16 & 55.54 & 62.12 & -12.03 & 4.61 & $\sim$240 & 76.01 & 64.35 & 78.02 & -13.99 & 13.90 \\
HiDe-LLaVA$^{\dagger}$ & -- & 60.77 & 55.31 & 60.68 & -6.82 & 3.83 & -- & 69.96 & 62.29 & 77.32 & -9.20 & 6.84 \\
SEFE$^{\dagger}$ & $\sim$80 & 65.01 & 58.51 & 63.63 & -8.13 & 1.01 & $\sim$40 & 75.98 & 66.54 & 78.76 & -11.33 & 9.97 \\
DISCO$^{\dagger}$ & $\sim$406 & 64.78 & 60.33 & 63.93 & -5.57 & 2.51 & $\sim$248 & 75.87 & \textbf{69.66} & \textbf{81.60} & -7.45 & 3.54 \\
\midrule
\multicolumn{13}{l}{\textit{Standalone (same routing)}}\\
LoRA-MoE & 164.5 & \textbf{67.48} & 48.03 & 54.16 & -24.32 & 25.98 & $\sim$246.7 & 69.53 & 46.06 & 60.31 & -28.16 & 42.13 \\
Monarch-MoE & 164.5 & 63.77 & 51.21 & 59.90 & -15.70 & 3.01 & $\sim$126.8 & 69.26 & 53.16 & 70.07 & -19.33 & 35.50 \\
\midrule
\multicolumn{13}{l}{\textit{Drop-in Monarch expert (same CL strategy and routing)}}\\
CL-MoE + Monarch & 105.6 & 64.45 & 53.88 & 62.08 & -13.21 & 1.53 & $\sim$127 & 68.39 & 56.30 & 72.79 & -14.51 & 19.23 \\
O-LoRA + Monarch & 99.9 & 64.48 & 54.08 & 61.61 & -12.99 & 1.80 & $\sim$120 & 67.91 & 57.83 & 73.27 & -12.11 & 7.66 \\
DISCO + Monarch & $\sim$102 & 63.89 & \textbf{63.56} & \textbf{64.86} & \textbf{-0.41} & \textbf{0.52} & $\sim$122 & 68.57 & 67.58 & 78.89 & \textbf{-1.18} & \textbf{0.10} \\
\bottomrule
\end{tabular}
\endgroup
\end{table*}

\begin{figure}[t]
\centering
\includegraphics[width=\linewidth]{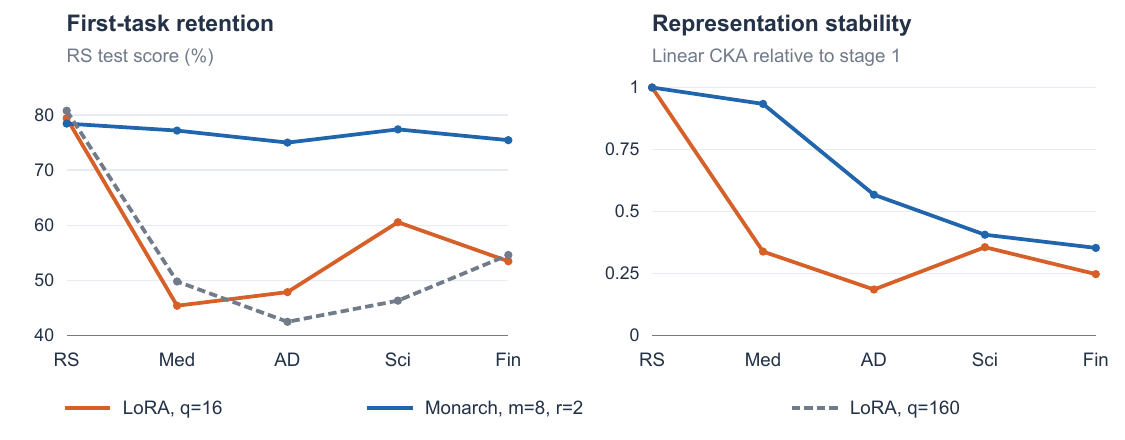}
\caption{\textbf{Retention and representation stability on DCL.} Left: first-task RS scores after each training stage. Raising dense LoRA rank from 16 to 160 does not recover Monarch's retention. Right: adapter-output linear CKA relative to the first checkpoint, using the same 256 RS examples at layer 16. Higher similarity indicates more stable representations at this measurement point; it does not by itself establish task isolation.}
\label{fig:results}
\end{figure}

\textbf{Standalone experts.}
At the equal DCL parameter budget, Monarch reduces first-task forgetting from 25.98 to 3.01\,pp and improves MFN from 48.03 to 51.21 and MAA from 54.16 to 59.90. Figure~\ref{fig:results} (left) shows that both models initially learn RS well, but their retained scores diverge during later tasks. Monarch's MFT is 3.71 points lower, so the retention gain accompanies weaker immediate learning. On UCIT, standalone Monarch improves MFN from 46.06 to 53.16 and BWT from $-28.16$ to $-19.33$ with fewer parameters, although first-task forgetting remains substantial at 35.50\,pp.

\textbf{Integration with continual-learning methods.}
On DCL, DISCO + Monarch achieves the best MFN, MAA, BWT, and first-task forgetting among the compared methods. MFN rises from 60.33 to 63.56 and first-task forgetting falls from 2.51 to 0.52\,pp, using approximately 102M rather than 406M adapter parameters. This combination improves retained performance while reducing the configured budget by about 75\%. The replacements in CL-MoE and O-LoRA also reduce first-task forgetting, from 7.74 to 1.53\,pp and from 4.61 to 1.80\,pp, at roughly one quarter of their budgets. Their MFN decreases by 1.00 and 1.46 points, however, and O-LoRA's BWT is slightly worse.

\textbf{Acquisition and retention.}
On UCIT, all three host replacements improve BWT and first-task forgetting, but lower MFN. DISCO + Monarch has the best BWT ($-1.18$) and first-task forgetting (0.10), while MFN falls from 69.66 to 67.58. Final performance reflects both immediate learning and subsequent retention: improved BWT outweighs the MFT decrease for DCL DISCO, but not for UCIT DISCO (Appendix~\ref{app:metrics}). The largest standalone RS separation occurs at the RS$\rightarrow$Med transition, motivating the repeated-prefix study below; later score recovery makes forgetting non-monotonic (Appendix~\ref{app:matrices}).

\subsection{Capacity Controls and Robustness}
\begin{table}[t]
\centering\small
\caption{\textbf{DCL capacity controls.} $\rho$ is the per-expert rank bound. The 44.5M variant matches the dense rank bound, whereas the 164.5M variant matches parameters. Output scales differ (Appendix~\ref{app:config}).}
\label{tab:capacity}
\begin{tabular}{lrrrrrrr}
\toprule
Expert & $\rho$ & Params (M) & MFT & MFN & MAA & BWT & $\rsf$ \\
\midrule
LoRA $q=16$ & 16 & 164.5 & 67.48 & 48.03 & 54.16 & -24.32 & 25.98 \\
LoRA $q=160$ & 160 & 1603.6 & 69.16 & 47.27 & 53.67 & -27.37 & 26.19 \\
Monarch $m=8,r=2$ & 128 & 164.5 & 63.77 & 51.21 & 59.90 & -15.70 & 3.01 \\
Monarch $m=4,r=1$ & 16 & 44.5 & 64.05 & 51.23 & 60.29 & -16.02 & 3.07 \\
\bottomrule
\end{tabular}
\end{table}

\textbf{Capacity controls.}
The equal-budget comparison leaves open whether Monarch's higher rank bound explains its retention. Table~\ref{tab:capacity} therefore includes a 44.5M Monarch with $(m,r)=(4,1)$, matching the per-expert rank bound of LoRA $q=16$. It retains first-task forgetting of 3.07\,pp and MFN 51.23. Conversely, increasing dense LoRA to $q=160$ uses 1.604B parameters and raises MFT to 69.16, but yields first-task forgetting of 26.19\,pp and MFN 47.27. Thus the observed retention advantage persists at the lower Monarch rank bound, while substantially increasing dense capacity does not recover it. These controls challenge a capacity-only explanation, although output scales and initialization dynamics are not fully matched (Appendix~\ref{app:config}).

\textbf{Repeatability across seeds.}
We repeat the RS$\rightarrow$Med prefix with seeds 42, 1234, and 2026 (Figure~\ref{fig:diagnostics}, left). First-task forgetting is $32.47\pm1.43$\,pp for LoRA and $1.12\pm0.13$\,pp for Monarch (mean $\pm$ sample standard deviation, $n=3$). All three seeds favor Monarch, with a mean gap of 31.35\,pp. This supports repeatability for the evaluated two-task setting; it does not constitute repeated five-task sequences. Appendix~\ref{app:seeds} gives the seedwise results, including seed 42, which is the prefix of the main run.

\textbf{Initialization and scale.}
The retention advantage is already present with zero initialization. For the 44.5M Monarch, SVD-Low further improves MFN from 51.23 to 52.05 and forgetting from 3.07 to 1.58\,pp, whereas SVD-High gives MFN 24.96 and forgetting 20.47 (Appendix~\ref{app:svd}). Scale sensitivity is also bounded: UCIT DISCO + Monarch has MFN 67.34--67.81 over $\alpha\in\{192,256,384,512\}$, but deteriorates at 16 and 1024. The complete sweeps retain these adverse settings and cases of low forgetting with weak acquisition; independent validation-based scale selection is not documented (Appendix~\ref{app:scaling}).

\subsection{Structural and Interpretability Analyses}
\label{sec:interpretability}
\begin{table}[t]
\centering\small
\caption{\textbf{Two-task block-count probe at fixed factor budget.} $mr=8$, $K=4$, $\alpha=8$ and scale $s=1$. Increasing $m$ changes the rank bound; each task is trained for one epoch.}
\label{tab:blocks}
\begin{tabular}{lrrrrrr}
\toprule
$m$ & $r$ & $\rho=m^2r$ & RS immediate & RS retained & Med & $\rsf$ \\
\midrule
1 & 8 & 8 & 79.35 & 76.06 & 52.91 & 3.29 \\
2 & 4 & 16 & 79.17 & 76.07 & 52.03 & 3.10 \\
4 & 2 & 32 & 78.95 & 76.63 & 51.45 & 2.32 \\
8 & 1 & 64 & 78.77 & 77.03 & 50.48 & 1.74 \\
\bottomrule
\end{tabular}
\end{table}

\textbf{Partition structure and retention.}
Section~\ref{sec:local_support} identifies local output support as the structural property motivating Monarch-MoE. To examine its relation to learning behavior, we vary the symmetric block count while fixing $mr=8$, $K=4$, and $\alpha=8$ (Table~\ref{tab:blocks}). The factor budget and output scale remain fixed, while the rank bound increases from 8 to 64. First-task forgetting decreases from 3.29 to 1.74\,pp, alongside a decrease in immediate Med accuracy from 52.91 to 50.48. Finer partitions therefore coincide with better retention and weaker acquisition in this probe, which trains each task for one epoch rather than using the full DCL schedule.

\begin{figure}[t]
\centering
\includegraphics[width=\linewidth]{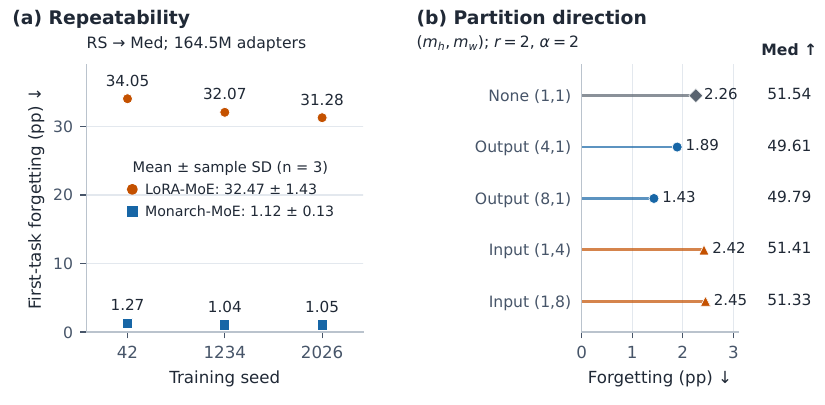}
\caption{\textbf{Repeatability and directional partition diagnostics.} Left: all three seeds for the equal-budget RS$\rightarrow$Med comparison, with mean $\pm$ sample standard deviation ($n=3$); RS and Med use one and three epochs. Right: complete directional probe with $r=2$ and $\alpha=2$, one epoch per task, and one run per configuration. Points show first-task forgetting; the Med column reports immediate new-task scores. Factor allocation and effective scale vary across partitions. The panels use different protocols and axis ranges and should not be compared as one experiment.}
\label{fig:diagnostics}
\end{figure}

A directional probe with $r=2$ and $\alpha=2$ further examines output-local support (Figure~\ref{fig:diagnostics}, right). Relative to 2.26\,pp forgetting without partitioning, output-only blocking gives 1.43 at $(m_h,m_w)=(8,1)$, whereas input-only blocking gives 2.45 at $(1,8)$. This pattern is consistent with the proposed role of output-local updates. Partition direction also changes factor allocation and normalization, however, so it does not isolate that property from scale (Appendix~\ref{app:blocks}).

\textbf{Geometry of learned updates.}
We next examine whether successive tasks use less-overlapping update directions. The archived analysis forms per-task expert weight increments, including the permutation in $LPR$, and compares the subspaces spanned by their top eight left singular directions. The reported principal angles are consistently larger for Monarch across adjacent-task transitions, placing these learned update subspaces closer to orthogonality than those of LoRA. This observation supports the hypothesis that the structure favors less-aligned task updates; it is not a measurement of raw task gradients or a guarantee of zero interference. Appendix~\ref{app:mechanism} reports the numerical summaries and their provenance limits: the underlying arrays and aggregation details are not fully retained.

\textbf{Stability of old-task representations.}
To relate update behavior to the representations used on an earlier task, we measure adapter-output linear CKA~\citep{kornblith2019similarity} at layer 16 on the same 256 RS examples, taking the first checkpoint as reference. After RS$\rightarrow$Med, CKA is 0.934 for Monarch and 0.338 for LoRA. Monarch remains higher at each subsequent checkpoint (Figure~\ref{fig:results}, right); LoRA also exhibits an intermediate recovery. These measurements complement the retained-score trajectories: Monarch preserves both greater similarity at this representation site and higher RS performance. They describe one layer and sample set, rather than establishing preservation throughout the network.

\textbf{Expert utilization.}
Finally, both methods have nearly uniform aggregate expert-selection shares across all five DCL stages (23.05--26.80\%, with normalized entropy at least 0.9991). Gross load imbalance is therefore an unlikely explanation for the retention gap, although these summaries do not establish identical token-level routing. Together, the partition probes, update subspaces, and representation measurements support the proposed interference interpretation, while the unmatched optimization choices prevent a complete causal attribution to structure alone.

\subsection{Single-GPU Training and Inference}
\label{sec:systems}
\textbf{Training cost and feasibility.}
On one RTX~4090, a DCL-RS probe with microbatch 4 and accumulation 4 compares standalone models at the same 164.5M adapter budget. Monarch reaches 1.606 samples/s versus LoRA's 1.344, a 19.5\% increase, with maximum recorded peaks of 19.300 versus 20.216\,GB. This measures implementation cost at equal parameter count. Separately, a compact five-expert Monarch configuration completes at 23,410\,MB and 1.039 samples/s; the local dense comparison exits abnormally at 23,966\,MB without an identified failure cause. These short probes establish feasibility for the tested configurations, rather than full-sequence timing (Appendix~\ref{app:efficiency}).

\textbf{Generation cost.}
An archived generation benchmark uses model-level compilation on 500 RS examples. The 44.5M Monarch reaches 5.42 samples/s versus 5.90 for the 164.5M LoRA model, an 8.1\% throughput reduction with fewer adapter parameters. Appendix~\ref{app:efficiency} separately reports forward-probe timings, which do not measure autoregressive generation. The results support single-GPU training and inference with workload-dependent costs; they do not isolate the contribution of individual execution optimizations.

% Keep the summary together below the final diagnostic figure.
\newpage
% CONCLUSION-D01 / v015 / R025: author-requested summary, pending review.
\section{Conclusion and Limitations}
We presented Monarch-MoE for multimodal continual instruction tuning, using expert structure to limit the direct spread of individual block-pair updates. Block-diagonal factors and a fixed permutation combine local output support with connections between all input and output groups. In the main comparisons, structured experts improve first-task retention; capacity controls, learned update subspaces, and old-task representation measurements provide complementary support for the proposed interference interpretation. Compact factors also reduce adapter storage, while the optimized implementation supports fine-tuning and inference on a single RTX~4090 in the measured configurations. These results support expert parameterization as a useful design choice for continual adaptation under memory constraints.

The gains involve method-dependent acquisition--retention trade-offs and do not guarantee higher final performance. Our evaluation covers one backbone and fixed task orders, with multi-seed evidence limited to the RS$\rightarrow$Med prefix. Local output support does not guarantee task isolation, and unmatched effective update scales limit causal attribution. Full-sequence repetitions, scale-matched controls, and evaluations on broader backbones and longer sequences remain necessary to establish the reach of the approach.

\label{main-end}
\clearpage
\subsection*{Reproducibility Statement}
The exact factor layout and scaling convention are given in Section~\ref{sec:method}; Appendix~\ref{app:rank} derives parameter and rank bounds. Appendices~\ref{app:config}--\ref{app:matrices} provide configuration distinctions, metric formulas, and stagewise results, including the scope of repeated runs. Appendix~\ref{app:provenance} identifies measurements retained only as analysis summaries and distinguishes configuration-based parameter estimates from checkpoint counts. These details support reproduction without implying that every original checkpoint or diagnostic log is included in this manuscript.

\subsection*{AI Use Statement}
Generative AI tools assisted with manuscript drafting and revision, literature retrieval and reference checking, research framing and experimental analysis, mathematical derivations and proof arguments, and preparation of tables, figures, and LaTeX source. Numerical results in this manuscript are drawn from recorded experiments or explicitly computed from their task-score matrices; the revision did not generate new training results. The authors are responsible for verifying the final claims, references, and artifacts.

\subsection*{Ethics Statement}
This work studies adaptation of an existing multimodal model on established benchmarks. Improved retention does not establish safety, fairness, or suitability for medical or autonomous-driving deployment. The method inherits limitations and potential biases of the backbone and datasets. Lower adaptation costs may broaden both beneficial use and misuse; task scores alone do not resolve these risks.

\bibliography{references}
\bibliographystyle{iclr2027_conference}
\clearpage
\appendix
\section{Structure, Rank, and Local Update Support}
\subsection{Parameter count and attainable rank}
\label{app:rank}
There are $m_hm_w$ block pairs. Each pair contains $hr$ parameters in $L_{i,j}$ and $rw$ in $R_{j,i}$. Summing gives
\[
 m_hm_wr(h+w)=r(m_wd_o+m_hd_i).
\]
The intermediate space contains $\rho=m_hm_wr$ channels. Since $P$ is a permutation, $\operatorname{rank}(LPR)\leq\min(d_o,d_i,\rho)$. If $h\geq m_wr$ and $w\geq m_hr$, choose each output block of $L$ to have full column rank and each input block of $R$ to have full row rank. Then $L$ is injective on the $\rho$-dimensional intermediate space, $R$ is surjective onto it, and $P$ is invertible. Hence $\operatorname{rank}(LPR)=\rho$. For a square symmetric partition, these conditions reduce to $m^2r\leq d$.

At a factor budget $2dmr$, a square LoRA expert has rank at most $q=mr$ and Monarch has attainable rank $m^2r$ under the preceding dimensional condition. Each \emph{block pair} of Monarch nevertheless has rank at most $r$. The two matrix families therefore impose different constraints; the rank comparison is not a containment or universal approximation claim. Routers add parameters outside this derivation, and a mixture's effective rank is not necessarily the rank of one expert.

\subsection{Jacobian support and the permutation}
\label{app:gradient}
Let $E_i$ embed an $h$-dimensional vector into output slice $i$, and let $S_j$ select input slice $j$. The expert matrix is
\[
 M=\sum_{i=1}^{m_h}\sum_{j=1}^{m_w}E_iL_{i,j}R_{j,i}S_j.
\]
For arbitrary changes $\Delta L_{i,j},\Delta R_{j,i}$ to just one pair,
\[
 \Delta M=E_i\bigl(\Delta L_{i,j}R_{j,i}+L_{i,j}\Delta R_{j,i}
       +\Delta L_{i,j}\Delta R_{j,i}\bigr)S_j.
\]
Thus $E_{i'}^\top\Delta M=0$ for $i'\ne i$. This finite-update statement implies the local gradient support in Eq.~\eqref{eq:grad}. It holds for one pair, not for an entire stored $R$ block (which includes channels to multiple output slices), a full expert update, or the end-to-end network. Gate derivatives and subsequent dense layers can introduce further coupling. A forgetting bound would additionally require assumptions on task gradients, routing, feature distributions, and layer propagation that are not established here.

For $m_h=m_w=m$, both block-diagonal factors have $m$ groups with $mr$ intermediate coordinates per group. Setting $P=I$ connects only corresponding groups, producing a block-diagonal map whose rank can still reach $m^2r$ when dimensions permit. The interleaving permutation instead supplies rank-$r$ channels to every input--output block pair. Consequently, off-diagonal zeros in $L$ and $R$ do not imply off-diagonal zeros in $LPR$. We make no claim that removing $P$ must reduce the global rank by a factor of $m$.

\subsection{Blockwise SVD initialization}
\label{app:svd}
For pretrained block $W_{ij}=U_{ij}\Sigma_{ij}V_{ij}^{\top}$, the optional initialization assigns disjoint groups of $r$ singular components to each of $K$ experts, up to the available block rank. SVD-High chooses the largest $Kr$ singular values and SVD-Low chooses the smallest. For a selected group $\mathcal I_e$ and positive compensation $c$, the implementation uses
\[
 L^{(e)}_{i,j}=U_{ij,\mathcal I_e}(\Sigma_{ij,\mathcal I_e}/c)^{1/2},\qquad
 R^{(e)}_{j,i}=(\Sigma_{ij,\mathcal I_e}/c)^{1/2}V_{ij,\mathcal I_e}^{\top}.
\]
The frozen residual contains the unassigned singular components. This gives the algebraic reconstruction $W=W_{\mathrm{res}}+c\sum_eL_ePR_e$ up to numerical precision. The actual network instead uses weighted, potentially sparse gates. Unless their coefficients match the reconstruction coefficients, this identity does not imply that the initialized network exactly reproduces the pretrained function. The experiment tests the resulting initialization, including its routing interaction.
\begin{table}[t]
\centering\small
\caption{\textbf{Blockwise SVD initialization, DCL.} Standalone $K=4,m=4,r=1,\alpha=1$; the same 44.5M adapter budget.}
\label{tab:svd}
\begin{tabular}{lrrrrr}
\toprule
Initialization & MFT & MFN & MAA & BWT & $\rsf$ \\
\midrule
Zero & 64.05 & 51.23 & 60.29 & -16.02 & 3.07 \\
SVD-Low & 62.22 & 52.05 & 59.12 & -12.71 & 1.58 \\
SVD-High & 45.62 & 24.96 & 30.01 & -25.82 & 20.47 \\
\bottomrule
\end{tabular}
\end{table}

\section{Experimental Configuration and Metrics}
\subsection{Configuration distinctions and parameter accounting}
\label{app:config}
The backbone is LLaVA-1.5-7B. MLLM-DCL follows the domain order remote sensing (RS), medicine (Med), autonomous driving (AD), science (Sci), and finance (Fin). UCIT follows ImageNet-R, ArxivQA, VizWiz-Caption, IconQA, CLEVR-Math, and Flickr30k. DCL standalone mixtures use $K=4$ experts and host replacements use $K=5$; UCIT uses $K=6$. Adaptation targets the query, key, value, and output attention projections and the up, down, and gate MLP projections in 32 transformer layers (224 adapted linear maps). For the LLaMA-7B dimensions, the sum of input plus output widths across these maps is 2,498,560. A dense rank-$q$ collection with $K$ experts therefore has $2{,}498{,}560Kq$ factor parameters; a symmetric Monarch has $2{,}498{,}560Kmr$. The counted standalone router adds $1{,}138{,}688K$ parameters. Counts exclude the frozen model, projector, and other non-adapter tensors; they are not the total size of a saved checkpoint.

\begin{table}[t]
\centering\small
\caption{Main expert configurations. $q$ is a per-expert LoRA rank; $(m,r)$ specifies symmetric Monarch. $\alpha$ is the adapter's implemented multiplier before rank normalization. Counts marked C come from archived checkpoint-count summaries; E denotes an estimate from configurations.}
\label{tab:config}
\begin{tabular}{llrrrr}
\toprule
Setting & Expert & $K$ & Rank specification & $\alpha$ & Params (M)\\
\midrule
DCL standalone & LoRA &4&$q=16$&32&164.463 C\\
DCL capacity & LoRA &4&$q=160$&320$^{*}$&1603.633 C\\
DCL standalone & Monarch &4&$(8,2)$&1&164.463 C\\
DCL rank control & Monarch &4&$(4,1)$&1&44.532 C\\
DCL CL-MoE & Monarch &5&$(4,2)$&16&105.636 C\\
DCL O-LoRA & Monarch &5&$(4,2)$&16&99.942 C\\
DCL DISCO & Monarch &5&$(4,2)$&256&$\sim$102 E\\
UCIT standalone & LoRA &6&$q=16$&32&$\sim$246.694 E\\
UCIT standalone & Monarch &6&$(4,2)$&16&$\sim$126.763 E\\
UCIT CL-MoE & Monarch &6&$(4,2)$&64&$\sim$127 E\\
UCIT O-LoRA & Monarch &6&$(4,2)$&64&$\sim$120 E\\
UCIT DISCO & Monarch &6&$(4,2)$&256&$\sim$122 E\\
\bottomrule
\end{tabular}
\end{table}

\textbf{Native baseline budgets.} MCITlib's DCL total-rank setting 160 is divided over five task experts, yielding $q=32$, not 160 per expert. UCIT uses total rank 96 over six experts, yielding $q=16$. O-LoRA has no adapter router: its corresponding factor estimates are 399.77M and 239.86M. CL-MoE adds routing parameters, approximately 405.46M and 246.69M. MoELoRA uses total ranks 35 and 18, yielding about 93.14M and 51.81M including the counted router convention. DISCO additionally maintains routing/subspace state; its table counts are approximate configuration budgets rather than measured checkpoints. We omit a parameter estimate for HiDe-LLaVA because its layer-dependent allocation is not captured reliably by a uniform-rank formula. These estimates should not be interpreted as equal implementation-level peak-memory measurements.

\textbf{Training and dispatch.} Full DCL uses epochs $(1,3,1,2,1)$ and learning rate $2\times10^{-5}$; UCIT uses one epoch per task and $2\times10^{-4}$. Standalone mixtures use top-2 routing. CL-MoE, O-LoRA, and DISCO retain their respective task/instance dispatch and training strategies. The retained DCL standalone snapshots include four GPUs with microbatch 4 and accumulation 4 for LoRA and the initial Monarch stages, and microbatch 2 with accumulation 8 for later Monarch stages; both give effective batch 64. UCIT main-configuration snapshots use $4\times2\times8$ or $2\times2\times16$ (GPUs $\times$ microbatch $\times$ accumulation). Host-replacement snapshots include both of these batch-64 settings, while separate three-GPU DCL O-LoRA files specify $3\times2\times11=66$. These are configuration records, not complete launch logs uniquely identifying every reported run. GPU counts follow configuration fields rather than filenames. The separate short block probes use microbatch 4 and accumulation 4, one epoch on each of RS and Med. Efficiency measurements use their own single-GPU protocol; they are not full-sequence timing measurements.

\phantomsection
\label{app:integration}
\textbf{Host-specific regularization.}
The O-LoRA replacement adapts the host's orthogonality strategy to the structured factors. For input block $j$, let $\mathcal R_j^{(t)}\in\mathbb R^{m_hr\times w}$ stack the current expert's $R_{j,i}$ factors. The retained implementation computes the entrywise mean absolute value of $\mathcal R_j^{(t)}(\mathcal R_j^{(p)})^\top$ for each previous expert $p<t$, with previous factors detached from this penalty. It averages over input blocks and previous experts, then over adapted layers, and adds the weighted term to the instruction-tuning loss. Thus the host's constraint is retained through a blockwise adaptation, not an unchanged dense-factor formula. Run-specific regularization coefficients are not established by this implementation alone. This auxiliary objective is distinct from the local support property of Eq.~\eqref{eq:localchange}, which follows from the factorization without an orthogonality penalty.

\textbf{Comparison limits.} DCL standalone scales differ: LoRA $q=16,\alpha=32$ has scale 2; Monarch $(8,2),\alpha=1$ has scale $1/16$; Monarch $(4,1),\alpha=1$ has scale $1/4$. The archived experiment outline reports $\alpha=320$ (scale 2) for high-rank LoRA; this multiplier is summary-reported ($^{*}$ in Table~\ref{tab:config}), since the corresponding detailed adapter configuration is not retained here. UCIT standalone LoRA and Monarch both have nominal scale 2, but different factor counts and rank bounds. Parameter- or rank-matching alone does not remove all optimization differences. The available records do not provide an independent validation-set protocol for the scale sweep.

\subsection{Metric definitions and rounding}
\label{app:metrics}
For $T$ tasks and the lower-triangular score matrix $a$, we compute
\begin{align}
 \mathrm{MFT}&=\frac{1}{T}\sum_{j=1}^{T}a_{j,j},&
 \mathrm{MFN}&=\frac{1}{T}\sum_{j=1}^{T}a_{T,j},\\
 \mathrm{MAA}&=\frac{1}{T}\sum_{t=1}^{T}\frac{1}{t}\sum_{j=1}^{t}a_{t,j},&
 \mathrm{BWT}&=\frac{1}{T-1}\sum_{j=1}^{T-1}(a_{T,j}-a_{j,j}),\\
 \rsf&=a_{1,1}-a_{T,1}.&&
\end{align}
MFT measures immediate learning, not zero-shot forward transfer. Negative $\rsf$ indicates improvement on task 1 relative to its first checkpoint, which can also occur when that checkpoint learned poorly. The four aggregate metrics and $\rsf$ capture different aspects of the sequence.

We recompute all displayed aggregates using the archived two-decimal task scores and round half up to two decimals. Differences of 0.01 from earlier reported aggregates can result from rounding convention or unavailable precision. Baseline matrices correspond to the MCITlib v3 protocol. For DCL O-LoRA, the archived matrix gives BWT $-12.03$, whereas the MCITlib v3 summary table reports $-12.53$; we use the reproducible matrix calculation and disclose this discrepancy. For the dense $q=160$ control, 47.27 is MFN; its MAA is 53.67. We do not treat a log field named \texttt{average\_accuracy} as MAA without checking its definition.

\textbf{Acquisition--retention decomposition.}
The metric definitions yield the exact identity
\begin{equation}
 \mathrm{MFN}=\mathrm{MFT}+\frac{T-1}{T}\,\mathrm{BWT}.
 \label{eq:tradeoff}
\end{equation}
Each final score equals its immediate score plus its subsequent change. For DCL DISCO, the BWT improvement contributes about 4.13 points, outweighing the 0.89-point MFT decrease. For UCIT DISCO, the corresponding 5.23-point contribution does not offset the 7.30-point MFT decrease. Displayed values differ slightly through rounding. Thus improved retention can coexist with lower final performance; first-task forgetting alone cannot describe this trade-off.

\section{Repeatability and Structural Ablations}
\subsection{Three-seed RS-to-Med check}
\label{app:seeds}
\begin{table}[h]
\centering\small
\caption{First-task forgetting (pp) over three seeds on the RS$\rightarrow$Med prefix. Both models use the 164.5M-parameter standalone configurations. The last column reports mean $\pm$ sample standard deviation ($n=3$); these are not repeated full DCL sequences.}
\label{tab:seeds}
\begin{tabular}{lrrrr}
\toprule
Expert & Seed 42 & Seed 1234 & Seed 2026 & Mean $\pm$ std\\
\midrule
LoRA $q=16$ &34.05&32.07&31.28&$32.47\pm1.43$\\
Monarch $(8,2)$ &1.27&1.04&1.05&$1.12\pm0.13$\\
\bottomrule
\end{tabular}
\end{table}
Monarch has lower forgetting in all three seeds, with a mean gap of 31.35\,pp. Sample standard deviations use the $n-1$ denominator. Seed 42 is the prefix of the main standalone run; it is not an additional independent repetition. Seeds 1234 and 2026 repeat the same two-task protocol and expert configurations. These measurements establish repeatability over the evaluated prefix, while full-sequence multi-seed evaluation remains outside the reported study.

\subsection{Block count and partition direction}
\label{app:blocks}
Table~\ref{tab:blocks} holds $mr=8$; under the stored factor shapes this is a fixed-budget sweep with varying rank bound $m^2r$, not a fixed-rank sweep. Table~\ref{tab:blocksfull} adds asymmetric partitions and a symmetric $r=1$ sweep. In the latter, factor count grows linearly with $m$, rank bound grows quadratically, and the scale is fixed at 1. Thus neither sweep simultaneously holds rank and parameters fixed.
\begin{table}[t]
\centering\small
\caption{\textbf{Directional and fixed-$r$ probes.} Upper: output-only versus input-only partitions. Lower: symmetric $r=1$ sweep, with increasing parameter count; it is not a same-budget experiment. The $(8,8,1)$ row is the same run as Table~\ref{tab:blocks}, not an independent replicate.}
\label{tab:blocksfull}
\begin{tabular}{lrrrrrrrr}
\toprule
$m_h$ & $m_w$ & $r$ & $\alpha$ & $s$ & RS immediate & RS retained & Med & $\rsf$ \\
\midrule
1 & 1 & 2 & 2 & 1.00 & 78.98 & 76.72 & 51.54 & 2.26 \\
4 & 1 & 2 & 2 & 0.25 & 78.70 & 76.81 & 49.61 & 1.89 \\
8 & 1 & 2 & 2 & 0.13 & 78.42 & 76.99 & 49.79 & 1.43 \\
1 & 4 & 2 & 2 & 1.00 & 78.95 & 76.53 & 51.41 & 2.42 \\
1 & 8 & 2 & 2 & 1.00 & 78.90 & 76.45 & 51.33 & 2.45 \\
\midrule
1 & 1 & 1 & 1 & 1.00 & 78.88 & 76.53 & 50.34 & 2.35 \\
2 & 2 & 1 & 2 & 1.00 & 78.84 & 76.75 & 50.55 & 2.09 \\
4 & 4 & 1 & 4 & 1.00 & 78.81 & 76.64 & 49.97 & 2.17 \\
8 & 8 & 1 & 8 & 1.00 & 78.77 & 77.03 & 50.48 & 1.74 \\
\bottomrule
\end{tabular}
\end{table}

For asymmetric partitions, both $r(m_wd_o+m_hd_i)$ and $\alpha/(m_hr)$ must be tracked. The $\alpha=2,r=2$ output-only runs have scales $1/4$ and $1/8$, whereas input-only runs retain scale 1. For non-square projections, exchanging $m_h$ and $m_w$ also changes parameter count. Lower forgetting in the output-only setting is compatible with localized output support but cannot isolate its causal effect from scale and allocation changes. A decisive control would match these quantities and repeat the runs across seeds.

\section{Representation and Routing Diagnostics}
\label{app:mechanism}
\textbf{Linear CKA.} The retained activation-analysis JSON evaluates 256 RS test examples selected with seed 0, at layer 16, across five checkpoints. Each checkpoint records 160,793 token representations with hidden dimension 4096. CKA compares centered adapter-output feature matrices against the first checkpoint, using $\|X^\top Y\|_F^2/(\|X^\top X\|_F\|Y^\top Y\|_F)$~\citep{kornblith2019similarity}. LoRA gives $(1,.3382,.1851,.3559,.2472)$ and Monarch gives $(1,.9339,.5675,.4062,.3530)$. These are the values plotted in Figure~\ref{fig:results}; no repeated-sample uncertainty is available.

\textbf{Output-subspace summaries.} The archived analysis constructs task increments $\Delta W_t$ from expert matrices (including $P$), extracts the top eight left singular directions, and computes principal angles. The reported adjacent transition summaries are LoRA $(45.4,51.1,44.9,49.6)^\circ$ and Monarch $(83.1,82.2,82.2,83.8)^\circ$. The archived overall values are $47.7^\circ$ and $82.8^\circ$; its across-layer/expert aggregation details and underlying arrays are not fully retained, so these values are reported as descriptive summaries rather than an independently reproduced statistic. Large angles describe learned update subspaces and do not establish disjoint task information or zero interference.

\begin{table}[t]
\centering\small
\caption{Archived router-utilization summary. Selection share is the percentage of top-2 expert selections. Entropy is normalized by $\log 4$. ``Dead'' is the reported percentage of unused layer--expert slots, not an expert-selection share.}
\label{tab:routing}
\begin{tabular}{llrrrrrr}
\toprule
Expert & Task & E0 & E1 & E2 & E3 & Entropy & Dead (\%)\\
\midrule
Monarch&RS &25.83&24.25&24.63&25.29&.9998&.22\\
&Med&25.36&26.08&24.01&24.56&.9996&.78\\
&AD &25.71&26.26&24.42&23.60&.9994&1.67\\
&Sci&25.84&25.80&24.71&23.65&.9995&1.45\\
&Fin&26.04&26.14&24.77&23.05&.9991&2.34\\
\midrule
LoRA&RS &25.89&25.78&23.81&24.52&.9996&.45\\
&Med&25.55&25.27&23.60&25.58&.9996&.45\\
&AD &26.12&24.95&24.00&24.93&.9997&.89\\
&Sci&26.31&25.26&24.16&24.27&.9996&.45\\
&Fin&26.80&25.04&23.47&24.69&.9992&.67\\
\bottomrule
\end{tabular}
\end{table}
\textbf{Routing protocol.} The analysis samples 256 test examples per task with seed 0, evaluates the corresponding task-stage checkpoint, and aggregates over 224 router modules with four experts and top-2 selection. Taskwise selection shares are close to uniform, while some individual layer--expert slots remain unused. This checks gross load imbalance only: no hypothesis test or token-level routing-equivalence comparison was retained.

\section{Scale Sensitivity and Efficiency}
\subsection{Complete scale sweeps}
\label{app:scaling}
Tables~\ref{tab:scalesdcl} and~\ref{tab:scalesucit} retain complete sequences for the indicated scale settings. The selected main-table settings follow the archived experiments, without evidence of independent validation-based selection. Partial early sweeps are not promoted to full-sequence results. A large positive BWT or negative first-task forgetting can coexist with weaker immediate learning; this is visible in the DCL O-LoRA $\alpha=64$ run. The UCIT DISCO plateau over $\alpha=192$--512 is specific to this protocol, and the deterioration at 1024 shows that scale robustness is bounded.
\begin{table}[t]
\centering\small
\caption{\textbf{DCL scale sensitivity.} Complete runs at each indicated $\alpha$, with $m=4,r=2$ and zero initialization. No separate validation-based scale selection is documented.}
\label{tab:scalesdcl}
\begin{tabular}{lrrrrr}
\toprule
Method / $\alpha$ & MFT & MFN & MAA & BWT & $\rsf$ \\
\midrule
O-LoRA $16$ & 64.48 & 54.08 & 61.61 & -12.99 & 1.80 \\
O-LoRA $64$ & 58.97 & 59.46 & 57.63 & 0.61 & -10.32 \\
DISCO $16$ & 58.84 & 56.30 & 58.44 & -3.17 & 7.90 \\
DISCO $192$ & 63.14 & 62.68 & 63.70 & -0.58 & 1.13 \\
DISCO $256$ & 63.89 & 63.56 & 64.86 & -0.41 & 0.52 \\
\bottomrule
\end{tabular}
\end{table}

\begin{table}[t]
\centering\small
\caption{\textbf{UCIT scale sensitivity.} Complete runs at each indicated $\alpha$, with $m=4,r=2$ and zero initialization. No separate validation-based scale selection is documented.}
\label{tab:scalesucit}
\begin{tabular}{lrrrrr}
\toprule
Method / $\alpha$ & MFT & MFN & MAA & BWT & $\rsf$ \\
\midrule
CL-MoE $16$ & 69.50 & 55.14 & 73.36 & -17.23 & 28.27 \\
CL-MoE $64$ & 68.39 & 56.30 & 72.79 & -14.51 & 19.23 \\
O-LoRA $16$ & 69.31 & 57.61 & 75.16 & -14.04 & 15.13 \\
O-LoRA $64$ & 67.91 & 57.83 & 73.27 & -12.11 & 7.66 \\
DISCO $16$ & 63.94 & 61.01 & 76.22 & -3.52 & 0.20 \\
DISCO $192$ & 68.30 & 67.34 & 78.58 & -1.15 & 0.13 \\
DISCO $256$ & 68.57 & 67.58 & 78.89 & -1.18 & 0.10 \\
DISCO $384$ & 68.79 & 67.79 & 78.87 & -1.21 & 0.23 \\
DISCO $512$ & 68.73 & 67.81 & 78.78 & -1.10 & -0.16 \\
DISCO $1024$ & 59.75 & 58.75 & 75.68 & -1.20 & 0.23 \\
\bottomrule
\end{tabular}
\end{table}

\subsection{Single-GPU execution measurements}
\label{app:efficiency}
\begin{table}[!htbp]
\centering\small
\caption{Single-RTX-4090 DCL-RS measurements from the retained efficiency registry. Both standalone models have 164.5M adapter parameters, microbatch 4, and accumulation 4. Training memory is the maximum recorded step peak; inference memory is the mean forward-probe peak. GB follows the registry's unit label. Forward probes are not autoregressive generation.}
\begin{tabular}{lrr}
\toprule
Measurement & LoRA-MoE & Monarch-MoE\\
\midrule
Steady training-step time (s) &11.907&9.964\\
Training throughput (samples/s)&1.344&1.606\\
Maximum training peak (recorded GB)&20.216&19.300\\
Inference forward time (s)&0.725&0.647\\
Inference mean peak (recorded GB)&16.205&16.186\\
\bottomrule
\end{tabular}
\end{table}
The training throughput gain is $1.605833834/1.343740040-1=19.5\%$. These measurements show practical execution costs at one operating point. They do not isolate any individual optimization: the registry does not provide a controlled ablation of batched execution, fusion, and custom backward. No repetitions or error bars are retained for this probe.

\begin{table}[!htbp]
\centering\small
\caption{Separate five-expert capacity probe on one RTX~4090 at microbatch 4 and sequence limit 2048. Peak-memory units follow the source field (MB). The dense run's exit code does not identify its failure cause.}
\begin{tabular}{lrrr}
\toprule
Setting & Samples/s & Peak (MB) & Exit code\\
\midrule
Local LoRA $q=32,\alpha=64,K=5$ &--&23,966&1\\
Monarch $m=4,r=2,\alpha=16,K=5$ &1.039&23,410&0\\
\bottomrule
\end{tabular}
\end{table}
The local dense path's archived status is ``OOM\_or\_crash''; the detailed stderr/CUDA traceback is missing. The Monarch path completes. This demonstrates successful single-GPU execution of the compact configuration and a failure of the compared local path, not that all native CL-MoE or O-LoRA configurations exceed 24\,GB. Older labels that do not match their actual adapter configurations are excluded from native-method runtime comparisons.

\textbf{Autoregressive inference.}
A separate archived benchmark uses 500 RS test examples and model-level \texttt{torch.compile} with \texttt{max-autotune-no-cudagraphs}. Table~\ref{tab:generation} reports the same compilation level for both models. The log reports about three generated tokens per example, so this comparison concerns short-answer generation. The smaller Monarch has 8.1\% lower throughput and 230\,MiB lower peak memory. Layer-level kernel improvements alone had not improved generation consistently, motivating compilation beyond the expert operator. The retained experiment log contains these summaries; per-example timing traces and repeated-run uncertainty are unavailable. These results should not be combined with the forward-probe timings above as if they were one protocol.
\begin{table}[!htbp]
\centering\small
\caption{Archived single-GPU generation benchmark with model-level compilation. Monarch uses $m=4,r=1,K=4$; LoRA uses $q=16,K=4$. Peak memory is recorded in MiB.}
\label{tab:generation}
\begin{tabular}{lrrrr}
\toprule
Setting & Adapter (M) & Samples/s & Latency (ms) & Peak (MiB)\\
\midrule
LoRA-MoE &164.5&5.90&169.5&14,582\\
Monarch-MoE &44.5&5.42&184.5&14,352\\
\bottomrule
\end{tabular}
\end{table}

\textbf{Implementation and shape dependence.}
Factors are consolidated across experts with shapes $(K,m_w,m_hr,w)$ for $R$ and $(K,m_h,h,m_wr)$ for $L$. A five-dimensional contraction produces channels directly in $(i,j,a)$ order, avoiding a separate permutation copy. The training path saves the input and factors and recomputes the intermediate channels during backward. Batched matrix multiplications compute the output-factor and channel gradients; subsequent contractions compute the input-factor and input gradients. This avoids retaining the forward channel tensor across the backward boundary, but does not eliminate other activations or temporary backward buffers.

For supported shapes, the Triton forward kernel fuses the two factors and the permutation in one launch, encoding channel reordering in its indices and keeping intermediate channels on chip. The implementation can perform redundant input-factor products to use matrix-multiply instructions, selecting the required rows afterward. This trades arithmetic for reduced intermediate memory traffic. The fused-forward autograd path also recomputes channels in backward; it is distinct from the batched training path. The $m=4,r=1$ configuration uses the batched/compiled path rather than the fused path requiring a larger intermediate tile. The development record includes slower fused generation runs and does not establish a universal fused-kernel speedup. Model-level compilation addresses dispatch overhead beyond a single expert operator.

Per expert and token, the unfused factor arithmetic requires $r(m_wd_o+m_hd_i)$ multiply--accumulate operations, excluding dispatch. At equal factor counts, dense LoRA and Monarch have the same leading arithmetic count. The batched top-2 path can compute all expert outputs before selecting and weighting them; sparse gates therefore do not guarantee that unselected expert arithmetic is skipped. End-to-end memory and runtime depend on the chosen path, tensor shapes, compilation, and workload, as distinguished in the measurements above.

\textbf{Analytical optimizer-state savings.}
Reducing factors from roughly 400M to 100M saves 300M parameters. Two FP32 Adam moment arrays alone then save $8\times300\mathrm{M}=2.4$ decimal GB (about 2.24 GiB). This excludes weights, gradients, master copies, activations, temporary buffers, frozen-backbone memory, and optimizer sharding. It is an illustrative state calculation, not measured end-to-end GPU memory.

\section{Evidence Provenance and Scope}
\label{app:provenance}
The paper consolidates archived experiments, without new training runs. Main and capacity scores are recomputed from retained task matrices; baselines follow the MCITlib registry. DCL adapter counts come from checkpoint-inventory summaries; UCIT counts use six-expert configuration snapshots. Verification against original checkpoints remains necessary before treating configuration estimates as measured counts.

Block probes and CKA have structured results; router shares and principal angles have analysis summaries. Three-seed results use task-score records and finalized summaries. Unretained block-update distributions and unsupported off-diagonal-energy constants are excluded. Counts, forgetting metrics, and initialization descriptions are checked against the stored factor layout.

\section{Stagewise Task Scores}
\label{app:matrices}
These matrices support the main comparisons, capacity controls, initialization ablation, and scale sweeps. Baselines follow the MCITlib archive; other entries are retained experimental scores. Dashes indicate tasks not yet evaluated. Aggregate metrics follow Appendix~\ref{app:metrics}.
\clearpage
\subsection*{DCL task-score matrices}
Columns follow the task order in Appendix~\ref{app:config}; rows are successive training stages.\par\medskip
\noindent\begin{minipage}[t]{.48\linewidth}\centering\small\setlength{\tabcolsep}{3.5pt}
\textbf{LoRA-FT}\par\smallskip
\begin{tabular}{rrrrrr}\toprule
$t$ & 1 & 2 & 3 & 4 & 5\\
\midrule
1 & 78.32 & -- & -- & -- & --\\
2 & 74.68 & 57.53 & -- & -- & --\\
3 & 68.93 & 47.19 & 52.15 & -- & --\\
4 & 75.12 & 45.56 & 38.46 & 49.44 & --\\
5 & 69.65 & 41.59 & 25.43 & 40.88 & 87.45\\
\bottomrule\end{tabular}\end{minipage}\hfill
\noindent\begin{minipage}[t]{.48\linewidth}\centering\small\setlength{\tabcolsep}{3.5pt}
\textbf{O-LoRA}\par\smallskip
\begin{tabular}{rrrrrr}\toprule
$t$ & 1 & 2 & 3 & 4 & 5\\
\midrule
1 & 79.25 & -- & -- & -- & --\\
2 & 74.05 & 56.52 & -- & -- & --\\
3 & 76.06 & 43.71 & 52.32 & -- & --\\
4 & 76.60 & 44.87 & 40.57 & 50.58 & --\\
5 & 74.64 & 44.42 & 30.02 & 41.47 & 87.15\\
\bottomrule\end{tabular}\end{minipage}\par\bigskip\bigskip
\noindent\begin{minipage}[t]{.48\linewidth}\centering\small\setlength{\tabcolsep}{3.5pt}
\textbf{MoELoRA}\par\smallskip
\begin{tabular}{rrrrrr}\toprule
$t$ & 1 & 2 & 3 & 4 & 5\\
\midrule
1 & 79.09 & -- & -- & -- & --\\
2 & 74.78 & 58.73 & -- & -- & --\\
3 & 77.69 & 43.72 & 51.47 & -- & --\\
4 & 76.87 & 43.79 & 32.81 & 48.67 & --\\
5 & 77.54 & 41.85 & 27.62 & 40.13 & 86.75\\
\bottomrule\end{tabular}\end{minipage}\hfill
\noindent\begin{minipage}[t]{.48\linewidth}\centering\small\setlength{\tabcolsep}{3.5pt}
\textbf{CL-MoE}\par\smallskip
\begin{tabular}{rrrrrr}\toprule
$t$ & 1 & 2 & 3 & 4 & 5\\
\midrule
1 & 79.08 & -- & -- & -- & --\\
2 & 73.48 & 60.56 & -- & -- & --\\
3 & 72.61 & 44.42 & 51.62 & -- & --\\
4 & 71.02 & 48.04 & 37.70 & 50.28 & --\\
5 & 71.34 & 46.84 & 26.33 & 41.17 & 88.74\\
\bottomrule\end{tabular}\end{minipage}\par\bigskip\bigskip
\noindent\begin{minipage}[t]{.48\linewidth}\centering\small\setlength{\tabcolsep}{3.5pt}
\textbf{HiDe-LLaVA}\par\smallskip
\begin{tabular}{rrrrrr}\toprule
$t$ & 1 & 2 & 3 & 4 & 5\\
\midrule
1 & 78.14 & -- & -- & -- & --\\
2 & 74.26 & 58.05 & -- & -- & --\\
3 & 74.90 & 42.94 & 39.65 & -- & --\\
4 & 75.43 & 44.91 & 38.33 & 46.44 & --\\
5 & 74.31 & 48.95 & 33.21 & 38.54 & 81.55\\
\bottomrule\end{tabular}\end{minipage}\hfill
\noindent\begin{minipage}[t]{.48\linewidth}\centering\small\setlength{\tabcolsep}{3.5pt}
\textbf{SEFE}\par\smallskip
\begin{tabular}{rrrrrr}\toprule
$t$ & 1 & 2 & 3 & 4 & 5\\
\midrule
1 & 78.27 & -- & -- & -- & --\\
2 & 76.32 & 58.42 & -- & -- & --\\
3 & 77.22 & 49.13 & 52.49 & -- & --\\
4 & 77.83 & 47.70 & 43.01 & 49.04 & --\\
5 & 77.26 & 50.37 & 37.21 & 40.87 & 86.82\\
\bottomrule\end{tabular}\end{minipage}\par\bigskip\bigskip
\noindent\begin{minipage}[t]{.48\linewidth}\centering\small\setlength{\tabcolsep}{3.5pt}
\textbf{DISCO}\par\smallskip
\begin{tabular}{rrrrrr}\toprule
$t$ & 1 & 2 & 3 & 4 & 5\\
\midrule
1 & 79.00 & -- & -- & -- & --\\
2 & 77.35 & 52.77 & -- & -- & --\\
3 & 77.60 & 50.01 & 52.22 & -- & --\\
4 & 76.63 & 44.90 & 49.04 & 50.69 & --\\
5 & 76.49 & 44.48 & 44.84 & 46.61 & 89.22\\
\bottomrule\end{tabular}\end{minipage}\hfill
\noindent\begin{minipage}[t]{.48\linewidth}\centering\small\setlength{\tabcolsep}{3.5pt}
\textbf{LoRA-MoE}\par\smallskip
\begin{tabular}{rrrrrr}\toprule
$t$ & 1 & 2 & 3 & 4 & 5\\
\midrule
1 & 79.44 & -- & -- & -- & --\\
2 & 45.39 & 62.48 & -- & -- & --\\
3 & 47.86 & 36.27 & 53.16 & -- & --\\
4 & 60.55 & 34.30 & 27.32 & 52.43 & --\\
5 & 53.46 & 34.20 & 24.07 & 38.50 & 89.90\\
\bottomrule\end{tabular}\end{minipage}\par\bigskip\bigskip
\clearpage
\subsection*{DCL task-score matrices (continued)}
Columns follow the task order in Appendix~\ref{app:config}; rows are successive training stages.\par\medskip
\noindent\begin{minipage}[t]{.48\linewidth}\centering\small\setlength{\tabcolsep}{3.5pt}
\textbf{Monarch-MoE}\par\smallskip
\begin{tabular}{rrrrrr}\toprule
$t$ & 1 & 2 & 3 & 4 & 5\\
\midrule
1 & 78.45 & -- & -- & -- & --\\
2 & 77.18 & 54.54 & -- & -- & --\\
3 & 75.00 & 41.06 & 51.64 & -- & --\\
4 & 77.40 & 38.85 & 27.69 & 48.29 & --\\
5 & 75.44 & 35.06 & 20.32 & 39.31 & 85.94\\
\bottomrule\end{tabular}\end{minipage}\hfill
\noindent\begin{minipage}[t]{.48\linewidth}\centering\small\setlength{\tabcolsep}{3.5pt}
\textbf{CL-MoE + Monarch}\par\smallskip
\begin{tabular}{rrrrrr}\toprule
$t$ & 1 & 2 & 3 & 4 & 5\\
\midrule
1 & 79.12 & -- & -- & -- & --\\
2 & 78.23 & 56.83 & -- & -- & --\\
3 & 78.24 & 44.09 & 50.88 & -- & --\\
4 & 78.13 & 43.04 & 38.22 & 49.09 & --\\
5 & 77.59 & 39.73 & 25.28 & 40.48 & 86.34\\
\bottomrule\end{tabular}\end{minipage}\par\bigskip\bigskip
\noindent\begin{minipage}[t]{.48\linewidth}\centering\small\setlength{\tabcolsep}{3.5pt}
\textbf{O-LoRA + Monarch}\par\smallskip
\begin{tabular}{rrrrrr}\toprule
$t$ & 1 & 2 & 3 & 4 & 5\\
\midrule
1 & 78.97 & -- & -- & -- & --\\
2 & 78.40 & 56.23 & -- & -- & --\\
3 & 77.75 & 42.71 & 51.31 & -- & --\\
4 & 77.66 & 41.06 & 34.04 & 48.92 & --\\
5 & 77.17 & 38.73 & 26.68 & 40.88 & 86.95\\
\bottomrule\end{tabular}\end{minipage}\hfill
\noindent\begin{minipage}[t]{.48\linewidth}\centering\small\setlength{\tabcolsep}{3.5pt}
\textbf{DISCO + Monarch}\par\smallskip
\begin{tabular}{rrrrrr}\toprule
$t$ & 1 & 2 & 3 & 4 & 5\\
\midrule
1 & 78.84 & -- & -- & -- & --\\
2 & 78.62 & 49.79 & -- & -- & --\\
3 & 78.58 & 49.78 & 53.22 & -- & --\\
4 & 78.32 & 48.94 & 53.25 & 48.24 & --\\
5 & 78.32 & 48.79 & 53.24 & 48.09 & 89.37\\
\bottomrule\end{tabular}\end{minipage}\par\bigskip\bigskip
\noindent\begin{minipage}[t]{.48\linewidth}\centering\small\setlength{\tabcolsep}{3.5pt}
\textbf{O-LoRA $64$}\par\smallskip
\begin{tabular}{rrrrrr}\toprule
$t$ & 1 & 2 & 3 & 4 & 5\\
\midrule
1 & 60.97 & -- & -- & -- & --\\
2 & 72.56 & 44.55 & -- & -- & --\\
3 & 74.05 & 40.73 & 52.41 & -- & --\\
4 & 72.21 & 45.16 & 48.08 & 48.38 & --\\
5 & 71.29 & 43.88 & 46.57 & 46.99 & 88.56\\
\bottomrule\end{tabular}\end{minipage}\hfill
\noindent\begin{minipage}[t]{.48\linewidth}\centering\small\setlength{\tabcolsep}{3.5pt}
\textbf{DISCO $16$}\par\smallskip
\begin{tabular}{rrrrrr}\toprule
$t$ & 1 & 2 & 3 & 4 & 5\\
\midrule
1 & 72.95 & -- & -- & -- & --\\
2 & 73.25 & 41.87 & -- & -- & --\\
3 & 71.54 & 41.79 & 52.17 & -- & --\\
4 & 66.25 & 39.52 & 51.32 & 43.74 & --\\
5 & 65.05 & 39.10 & 50.82 & 43.08 & 83.45\\
\bottomrule\end{tabular}\end{minipage}\par\bigskip\bigskip
\noindent\begin{minipage}[t]{.48\linewidth}\centering\small\setlength{\tabcolsep}{3.5pt}
\textbf{DISCO $192$}\par\smallskip
\begin{tabular}{rrrrrr}\toprule
$t$ & 1 & 2 & 3 & 4 & 5\\
\midrule
1 & 77.51 & -- & -- & -- & --\\
2 & 77.40 & 48.05 & -- & -- & --\\
3 & 76.98 & 48.01 & 53.16 & -- & --\\
4 & 76.51 & 47.12 & 53.21 & 47.93 & --\\
5 & 76.38 & 47.00 & 53.18 & 47.77 & 89.05\\
\bottomrule\end{tabular}\end{minipage}\hfill
\noindent\begin{minipage}[t]{.48\linewidth}\centering\small\setlength{\tabcolsep}{3.5pt}
\textbf{LoRA $q=160$}\par\smallskip
\begin{tabular}{rrrrrr}\toprule
$t$ & 1 & 2 & 3 & 4 & 5\\
\midrule
1 & 80.79 & -- & -- & -- & --\\
2 & 49.79 & 64.34 & -- & -- & --\\
3 & 42.47 & 33.71 & 54.41 & -- & --\\
4 & 46.31 & 32.67 & 24.66 & 55.17 & --\\
5 & 54.60 & 31.69 & 20.94 & 38.01 & 91.11\\
\bottomrule\end{tabular}\end{minipage}\par\bigskip\bigskip
\clearpage
\subsection*{DCL task-score matrices (continued)}
Columns follow the task order in Appendix~\ref{app:config}; rows are successive training stages.\par\medskip
\noindent\begin{minipage}[t]{.48\linewidth}\centering\small\setlength{\tabcolsep}{3.5pt}
\textbf{Monarch $m=4,r=1$}\par\smallskip
\begin{tabular}{rrrrrr}\toprule
$t$ & 1 & 2 & 3 & 4 & 5\\
\midrule
1 & 78.54 & -- & -- & -- & --\\
2 & 77.35 & 54.98 & -- & -- & --\\
3 & 75.45 & 39.28 & 51.75 & -- & --\\
4 & 77.36 & 39.13 & 35.11 & 48.43 & --\\
5 & 75.47 & 34.10 & 20.64 & 39.40 & 86.54\\
\bottomrule\end{tabular}\end{minipage}\hfill
\noindent\begin{minipage}[t]{.48\linewidth}\centering\small\setlength{\tabcolsep}{3.5pt}
\textbf{SVD-Low}\par\smallskip
\begin{tabular}{rrrrrr}\toprule
$t$ & 1 & 2 & 3 & 4 & 5\\
\midrule
1 & 77.88 & -- & -- & -- & --\\
2 & 76.45 & 48.91 & -- & -- & --\\
3 & 72.28 & 37.99 & 52.03 & -- & --\\
4 & 77.07 & 37.08 & 34.94 & 46.41 & --\\
5 & 76.30 & 33.46 & 25.16 & 39.46 & 85.86\\
\bottomrule\end{tabular}\end{minipage}\par\bigskip\bigskip
\noindent\begin{minipage}[t]{.48\linewidth}\centering\small\setlength{\tabcolsep}{3.5pt}
\textbf{SVD-High}\par\smallskip
\begin{tabular}{rrrrrr}\toprule
$t$ & 1 & 2 & 3 & 4 & 5\\
\midrule
1 & 47.40 & -- & -- & -- & --\\
2 & 31.34 & 30.13 & -- & -- & --\\
3 & 36.63 & 25.74 & 48.09 & -- & --\\
4 & 16.65 & 0.01 & 0.79 & 23.12 & --\\
5 & 26.93 & 1.43 & 0.59 & 16.52 & 79.34\\
\bottomrule\end{tabular}\end{minipage}\hfill
\clearpage
\subsection*{UCIT task-score matrices}
Columns follow the task order in Appendix~\ref{app:config}; rows are successive training stages.\par\medskip
\noindent\begin{minipage}[t]{.48\linewidth}\centering\small\setlength{\tabcolsep}{3.5pt}
\textbf{LoRA-FT}\par\smallskip
\begin{tabular}{rrrrrrr}\toprule
$t$ & 1 & 2 & 3 & 4 & 5 & 6\\
\midrule
1 & 92.00 & -- & -- & -- & -- & --\\
2 & 90.63 & 91.03 & -- & -- & -- & --\\
3 & 74.00 & 91.40 & 61.90 & -- & -- & --\\
4 & 72.73 & 78.53 & 53.03 & 79.87 & -- & --\\
5 & 68.97 & 77.60 & 49.43 & 67.67 & 78.33 & --\\
6 & 58.03 & 77.63 & 44.39 & 67.40 & 61.77 & 58.22\\
\bottomrule\end{tabular}\end{minipage}\hfill
\noindent\begin{minipage}[t]{.48\linewidth}\centering\small\setlength{\tabcolsep}{3.5pt}
\textbf{O-LoRA}\par\smallskip
\begin{tabular}{rrrrrrr}\toprule
$t$ & 1 & 2 & 3 & 4 & 5 & 6\\
\midrule
1 & 91.40 & -- & -- & -- & -- & --\\
2 & 90.07 & 94.73 & -- & -- & -- & --\\
3 & 81.50 & 91.70 & 61.59 & -- & -- & --\\
4 & 82.40 & 77.73 & 54.06 & 76.63 & -- & --\\
5 & 81.07 & 77.87 & 50.62 & 62.00 & 73.53 & --\\
6 & 77.50 & 78.07 & 44.50 & 63.13 & 64.73 & 58.16\\
\bottomrule\end{tabular}\end{minipage}\par\bigskip\bigskip
\noindent\begin{minipage}[t]{.48\linewidth}\centering\small\setlength{\tabcolsep}{3.5pt}
\textbf{MoELoRA}\par\smallskip
\begin{tabular}{rrrrrrr}\toprule
$t$ & 1 & 2 & 3 & 4 & 5 & 6\\
\midrule
1 & 91.23 & -- & -- & -- & -- & --\\
2 & 90.07 & 92.80 & -- & -- & -- & --\\
3 & 80.30 & 92.03 & 60.22 & -- & -- & --\\
4 & 80.53 & 77.20 & 51.19 & 53.73 & -- & --\\
5 & 79.37 & 77.40 & 48.68 & 51.30 & 71.67 & --\\
6 & 70.07 & 77.70 & 44.69 & 50.03 & 54.03 & 57.34\\
\bottomrule\end{tabular}\end{minipage}\hfill
\noindent\begin{minipage}[t]{.48\linewidth}\centering\small\setlength{\tabcolsep}{3.5pt}
\textbf{CL-MoE}\par\smallskip
\begin{tabular}{rrrrrrr}\toprule
$t$ & 1 & 2 & 3 & 4 & 5 & 6\\
\midrule
1 & 91.30 & -- & -- & -- & -- & --\\
2 & 87.40 & 92.27 & -- & -- & -- & --\\
3 & 76.23 & 90.53 & 60.18 & -- & -- & --\\
4 & 77.33 & 76.20 & 51.09 & 54.33 & -- & --\\
5 & 75.73 & 75.70 & 48.27 & 52.80 & 73.23 & --\\
6 & 66.33 & 77.00 & 44.78 & 51.87 & 53.53 & 57.42\\
\bottomrule\end{tabular}\end{minipage}\par\bigskip\bigskip
\noindent\begin{minipage}[t]{.48\linewidth}\centering\small\setlength{\tabcolsep}{3.5pt}
\textbf{HiDe-LLaVA}\par\smallskip
\begin{tabular}{rrrrrrr}\toprule
$t$ & 1 & 2 & 3 & 4 & 5 & 6\\
\midrule
1 & 90.87 & -- & -- & -- & -- & --\\
2 & 88.83 & 92.27 & -- & -- & -- & --\\
3 & 88.17 & 91.90 & 54.46 & -- & -- & --\\
4 & 87.00 & 89.70 & 46.51 & 73.13 & -- & --\\
5 & 86.23 & 90.07 & 44.28 & 64.37 & 54.77 & --\\
6 & 84.03 & 90.73 & 44.43 & 58.93 & 41.37 & 54.25\\
\bottomrule\end{tabular}\end{minipage}\hfill
\noindent\begin{minipage}[t]{.48\linewidth}\centering\small\setlength{\tabcolsep}{3.5pt}
\textbf{SEFE}\par\smallskip
\begin{tabular}{rrrrrrr}\toprule
$t$ & 1 & 2 & 3 & 4 & 5 & 6\\
\midrule
1 & 90.80 & -- & -- & -- & -- & --\\
2 & 90.23 & 91.23 & -- & -- & -- & --\\
3 & 83.60 & 89.73 & 61.24 & -- & -- & --\\
4 & 84.17 & 77.20 & 56.36 & 78.93 & -- & --\\
5 & 82.90 & 77.13 & 53.54 & 71.43 & 75.73 & --\\
6 & 80.83 & 78.00 & 47.01 & 69.63 & 65.83 & 57.92\\
\bottomrule\end{tabular}\end{minipage}\par\bigskip\bigskip
\noindent\begin{minipage}[t]{.48\linewidth}\centering\small\setlength{\tabcolsep}{3.5pt}
\textbf{DISCO}\par\smallskip
\begin{tabular}{rrrrrrr}\toprule
$t$ & 1 & 2 & 3 & 4 & 5 & 6\\
\midrule
1 & 90.97 & -- & -- & -- & -- & --\\
2 & 90.07 & 91.77 & -- & -- & -- & --\\
3 & 89.40 & 92.53 & 61.39 & -- & -- & --\\
4 & 88.50 & 93.97 & 58.61 & 78.00 & -- & --\\
5 & 88.03 & 93.40 & 57.11 & 70.93 & 76.40 & --\\
6 & 87.43 & 93.07 & 46.96 & 68.13 & 65.70 & 56.69\\
\bottomrule\end{tabular}\end{minipage}\hfill
\noindent\begin{minipage}[t]{.48\linewidth}\centering\small\setlength{\tabcolsep}{3.5pt}
\textbf{LoRA-MoE}\par\smallskip
\begin{tabular}{rrrrrrr}\toprule
$t$ & 1 & 2 & 3 & 4 & 5 & 6\\
\midrule
1 & 91.13 & -- & -- & -- & -- & --\\
2 & 85.20 & 94.53 & -- & -- & -- & --\\
3 & 53.27 & 63.67 & 52.41 & -- & -- & --\\
4 & 62.77 & 27.33 & 26.66 & 54.70 & -- & --\\
5 & 44.83 & 13.00 & 3.90 & 42.73 & 72.87 & --\\
6 & 49.00 & 67.37 & 31.12 & 41.73 & 35.60 & 51.52\\
\bottomrule\end{tabular}\end{minipage}\par\bigskip\bigskip
\clearpage
\subsection*{UCIT task-score matrices (continued)}
Columns follow the task order in Appendix~\ref{app:config}; rows are successive training stages.\par\medskip
\noindent\begin{minipage}[t]{.48\linewidth}\centering\small\setlength{\tabcolsep}{3.5pt}
\textbf{Monarch-MoE}\par\smallskip
\begin{tabular}{rrrrrrr}\toprule
$t$ & 1 & 2 & 3 & 4 & 5 & 6\\
\midrule
1 & 91.33 & -- & -- & -- & -- & --\\
2 & 89.80 & 94.87 & -- & -- & -- & --\\
3 & 76.47 & 81.17 & 52.03 & -- & -- & --\\
4 & 77.37 & 77.07 & 39.91 & 54.13 & -- & --\\
5 & 73.47 & 32.07 & 28.99 & 51.87 & 71.67 & --\\
6 & 55.83 & 77.47 & 31.19 & 50.10 & 52.80 & 51.55\\
\bottomrule\end{tabular}\end{minipage}\hfill
\noindent\begin{minipage}[t]{.48\linewidth}\centering\small\setlength{\tabcolsep}{3.5pt}
\textbf{CL-MoE + Monarch}\par\smallskip
\begin{tabular}{rrrrrrr}\toprule
$t$ & 1 & 2 & 3 & 4 & 5 & 6\\
\midrule
1 & 90.03 & -- & -- & -- & -- & --\\
2 & 88.70 & 93.53 & -- & -- & -- & --\\
3 & 74.43 & 92.67 & 53.40 & -- & -- & --\\
4 & 78.97 & 77.37 & 43.69 & 51.90 & -- & --\\
5 & 76.90 & 77.40 & 40.09 & 50.67 & 68.93 & --\\
6 & 70.80 & 77.77 & 34.91 & 50.63 & 51.13 & 52.57\\
\bottomrule\end{tabular}\end{minipage}\par\bigskip\bigskip
\noindent\begin{minipage}[t]{.48\linewidth}\centering\small\setlength{\tabcolsep}{3.5pt}
\textbf{O-LoRA + Monarch}\par\smallskip
\begin{tabular}{rrrrrrr}\toprule
$t$ & 1 & 2 & 3 & 4 & 5 & 6\\
\midrule
1 & 89.93 & -- & -- & -- & -- & --\\
2 & 89.83 & 94.17 & -- & -- & -- & --\\
3 & 81.90 & 93.37 & 53.42 & -- & -- & --\\
4 & 84.93 & 57.57 & 51.44 & 50.73 & -- & --\\
5 & 84.60 & 59.83 & 49.33 & 52.10 & 66.60 & --\\
6 & 82.27 & 66.50 & 36.65 & 51.47 & 57.43 & 52.63\\
\bottomrule\end{tabular}\end{minipage}\hfill
\noindent\begin{minipage}[t]{.48\linewidth}\centering\small\setlength{\tabcolsep}{3.5pt}
\textbf{DISCO + Monarch}\par\smallskip
\begin{tabular}{rrrrrrr}\toprule
$t$ & 1 & 2 & 3 & 4 & 5 & 6\\
\midrule
1 & 90.53 & -- & -- & -- & -- & --\\
2 & 90.50 & 94.30 & -- & -- & -- & --\\
3 & 90.37 & 94.37 & 52.33 & -- & -- & --\\
4 & 90.43 & 94.33 & 51.52 & 52.40 & -- & --\\
5 & 90.43 & 94.30 & 51.32 & 52.27 & 69.77 & --\\
6 & 90.43 & 94.33 & 46.85 & 52.10 & 69.73 & 52.06\\
\bottomrule\end{tabular}\end{minipage}\par\bigskip\bigskip
\noindent\begin{minipage}[t]{.48\linewidth}\centering\small\setlength{\tabcolsep}{3.5pt}
\textbf{CL-MoE $16$}\par\smallskip
\begin{tabular}{rrrrrrr}\toprule
$t$ & 1 & 2 & 3 & 4 & 5 & 6\\
\midrule
1 & 90.40 & -- & -- & -- & -- & --\\
2 & 89.63 & 95.30 & -- & -- & -- & --\\
3 & 77.47 & 94.43 & 53.37 & -- & -- & --\\
4 & 77.70 & 78.20 & 43.86 & 53.67 & -- & --\\
5 & 75.20 & 78.50 & 40.53 & 52.40 & 72.00 & --\\
6 & 62.13 & 78.60 & 34.16 & 50.20 & 53.50 & 52.23\\
\bottomrule\end{tabular}\end{minipage}\hfill
\noindent\begin{minipage}[t]{.48\linewidth}\centering\small\setlength{\tabcolsep}{3.5pt}
\textbf{O-LoRA $16$}\par\smallskip
\begin{tabular}{rrrrrrr}\toprule
$t$ & 1 & 2 & 3 & 4 & 5 & 6\\
\midrule
1 & 90.33 & -- & -- & -- & -- & --\\
2 & 90.37 & 95.27 & -- & -- & -- & --\\
3 & 83.87 & 95.10 & 53.12 & -- & -- & --\\
4 & 83.53 & 80.23 & 47.66 & 53.63 & -- & --\\
5 & 82.40 & 80.63 & 47.13 & 51.57 & 71.20 & --\\
6 & 75.20 & 81.63 & 35.68 & 50.03 & 50.80 & 52.31\\
\bottomrule\end{tabular}\end{minipage}\par\bigskip\bigskip
\noindent\begin{minipage}[t]{.48\linewidth}\centering\small\setlength{\tabcolsep}{3.5pt}
\textbf{DISCO $16$}\par\smallskip
\begin{tabular}{rrrrrrr}\toprule
$t$ & 1 & 2 & 3 & 4 & 5 & 6\\
\midrule
1 & 88.97 & -- & -- & -- & -- & --\\
2 & 89.00 & 95.10 & -- & -- & -- & --\\
3 & 88.97 & 95.07 & 50.93 & -- & -- & --\\
4 & 88.90 & 94.77 & 50.68 & 48.13 & -- & --\\
5 & 88.93 & 94.70 & 50.40 & 46.93 & 50.67 & --\\
6 & 88.77 & 94.70 & 37.43 & 46.17 & 49.13 & 49.86\\
\bottomrule\end{tabular}\end{minipage}\hfill
\noindent\begin{minipage}[t]{.48\linewidth}\centering\small\setlength{\tabcolsep}{3.5pt}
\textbf{DISCO $192$}\par\smallskip
\begin{tabular}{rrrrrrr}\toprule
$t$ & 1 & 2 & 3 & 4 & 5 & 6\\
\midrule
1 & 90.10 & -- & -- & -- & -- & --\\
2 & 90.07 & 94.13 & -- & -- & -- & --\\
3 & 90.00 & 94.13 & 52.12 & -- & -- & --\\
4 & 90.13 & 94.47 & 51.34 & 51.83 & -- & --\\
5 & 90.17 & 94.47 & 50.58 & 51.43 & 69.60 & --\\
6 & 89.97 & 94.40 & 46.53 & 51.50 & 69.63 & 52.01\\
\bottomrule\end{tabular}\end{minipage}\par\bigskip\bigskip
\clearpage
\subsection*{UCIT task-score matrices (continued)}
Columns follow the task order in Appendix~\ref{app:config}; rows are successive training stages.\par\medskip
\noindent\begin{minipage}[t]{.48\linewidth}\centering\small\setlength{\tabcolsep}{3.5pt}
\textbf{DISCO $384$}\par\smallskip
\begin{tabular}{rrrrrrr}\toprule
$t$ & 1 & 2 & 3 & 4 & 5 & 6\\
\midrule
1 & 90.53 & -- & -- & -- & -- & --\\
2 & 90.47 & 94.07 & -- & -- & -- & --\\
3 & 90.43 & 94.07 & 52.11 & -- & -- & --\\
4 & 90.37 & 94.23 & 51.17 & 52.67 & -- & --\\
5 & 90.30 & 94.17 & 50.91 & 51.97 & 71.00 & --\\
6 & 90.30 & 94.17 & 46.68 & 52.13 & 71.07 & 52.36\\
\bottomrule\end{tabular}\end{minipage}\hfill
\noindent\begin{minipage}[t]{.48\linewidth}\centering\small\setlength{\tabcolsep}{3.5pt}
\textbf{DISCO $512$}\par\smallskip
\begin{tabular}{rrrrrrr}\toprule
$t$ & 1 & 2 & 3 & 4 & 5 & 6\\
\midrule
1 & 90.67 & -- & -- & -- & -- & --\\
2 & 90.67 & 92.87 & -- & -- & -- & --\\
3 & 90.73 & 92.80 & 52.26 & -- & -- & --\\
4 & 90.80 & 93.23 & 51.66 & 53.07 & -- & --\\
5 & 90.73 & 93.43 & 51.11 & 52.20 & 70.87 & --\\
6 & 90.83 & 93.40 & 47.03 & 51.97 & 71.00 & 52.61\\
\bottomrule\end{tabular}\end{minipage}\par\bigskip\bigskip
\noindent\begin{minipage}[t]{.48\linewidth}\centering\small\setlength{\tabcolsep}{3.5pt}
\textbf{DISCO $1024$}\par\smallskip
\begin{tabular}{rrrrrrr}\toprule
$t$ & 1 & 2 & 3 & 4 & 5 & 6\\
\midrule
1 & 91.13 & -- & -- & -- & -- & --\\
2 & 91.13 & 93.33 & -- & -- & -- & --\\
3 & 91.20 & 93.30 & 52.68 & -- & -- & --\\
4 & 91.13 & 93.50 & 51.82 & 52.70 & -- & --\\
5 & 91.10 & 93.67 & 51.17 & 51.47 & 15.73 & --\\
6 & 90.90 & 93.67 & 47.72 & 51.53 & 15.77 & 52.90\\
\bottomrule\end{tabular}\end{minipage}\hfill

\end{document}